\documentclass[11pt]{article}

\usepackage{amsmath}
\usepackage{graphicx}
\usepackage{titling}
\begin{document}

\title{\bf Benchmarking recognition results on word image datasets}

\author{Deepak Kumar, M N Anil Prasad and A G Ramakrishnan\\
\small Medical Intelligence and Language Engineering Laboratory\\
\small Department of Electrical Engineering\\
\small Indian Institute of Science \\
\small Bangalore, 560012, India\\
{\bf Email:} {deepak,anilprasadmn,ramkiag}@ee.iisc.ernet.in}

\maketitle

\begin{abstract}
{\it We have benchmarked the maximum obtainable recognition accuracy on various word image datasets using manual segmentation and a currently available commercial OCR. We have developed a Matlab program, with graphical user interface, for semi-automated pixel level segmentation of word images. We discuss the advantages of pixel level annotation. We have covered five databases adding up to over 3600 word images. These word images have been cropped from camera captured scene, born-digital and street view images. We recognize the segmented word image using the trial version of Nuance Omnipage OCR. We also discuss, how the degradations introduced during acquisition or inaccuracies introduced during creation of word images affect the recognition of the word present in the image. Word images for different kinds of degradations and correction for slant and curvy nature of words are also discussed. The word recognition rates obtained on ICDAR 2003, Sign evaluation, Street view, Born-digital and ICDAR 2011 datasets are 83.9\%, 89.3\%, 79.6\%, 88.5\% and 86.7\% respectively.}
\end{abstract}


\section{Introduction}
We have created pixel level annotation of word images publicly available for download, specifically for word image segmentation\footnote{Annotated Datasets: http://mile.ee.iisc.ernet.in/mile/download.html}. We have annotated different datasets consisting of different kinds of word images. To our knowledge, annotation at pixel level and among several datasets has not been carried out, until now. Small subsets from different datasets have been annotated and utilized for algorithms \cite{ananda, kasar, asif12}. We have annotated 3606 word images at pixel level. Annotation is not fully automated. Hence, it is a huge task as compared to similar tasks in computer vision or document imaging community. 

A human being requires a very short time to analyze any given image. To perform similar analysis by a computer algorithm is not simple. People analyze images using both top-down and bottom-up paradigms. Combining these two approaches is not an easy task. We often read that top-down is far better than bottom-up approach in image analysis. The relative contribution of top-down and bottom-up approaches in human vision is clearly unknown. An approach is developed to understand this contribution. For this approach, it was essential to annotate word images at the pixel level.  

We split the recognition of word from images into segmentation and recognition tasks. The term `binarization' is commonly used in place of segmentation. We require complex algorithms to segment an image. In document imaging community, conventional research primarily focused on digitization of scanned documents. It involved binarization of document image and recognition. In the section on annotation, we discuss known algorithms for segmentation of word images. These algorithms were helpful in improving the speed of pixel level annotation.

Annotated pixel level word images can be used to train and test any classifier. However, several good optical character recognition (OCR) engines are already available for Roman script \cite{abbyy, omnipage, tesseract, adobe}. Hence, we focus only on the annotation algorithm and annotating datasets. 

Necessity of annotation arises during benchmarking datasets. Earlier to our pixel level approach, scene-text images have largely been annotated using bounding box approach. It makes annotation an easier task. Of course few datasets do provide pixel level annotation, but they do not cover thousands of images.

The annotated images are passed on to the recognition stage. The recognition step can be performed using a training dataset or an OCR engine. We use the trial version of Omnipage Professional 16 OCR for recognition of characters in the binarized image \cite{omnipage} to create the benchmark recognition result. Definitely the numbers will slightly vary if we use any other standard OCR and hence the benchmark results we report here indicate a rough level of recognition that can be achieved, rather than the exact maximum value attainable in current circumstances.

If a single dataset is used in the experiments, it may lead to a dataset specific approach. So, to justify our approach that annotation is dataset independent, we cover five datasets for benchmarking. Either top-down or bottom-up approach is used in some datasets and both in others. These datasets are from ICDAR 2003 competition \cite{simon}, ICDAR 2011 competition \cite{karatzas, asif}, Street view \cite{wang10} and Sign evaluation datasets \cite{weinmann}. 

\section{Background}
When a camera captured image is presented to an OCR engine, the recognition performance is not necessarily very good. This led to spliting the process of word recognition in  camera captured images into two parts, namely localization (or detection) and recognition by Lucas et. al \cite{simon}. In International Conference on Document Analysis and Recognition (ICDAR) 2003, they organized separate competitions for text localization on camera captured images and recognition from the word images  extracted by placing a bounding box on the image.  They received five entries for text localization and none for word recognition. In the following ICDAR 2005 conference, text localization was the main theme and word recognition was skipped \cite{simon05}.

\begin{figure}[!b]
\centering
\includegraphics[width=3.6cm]{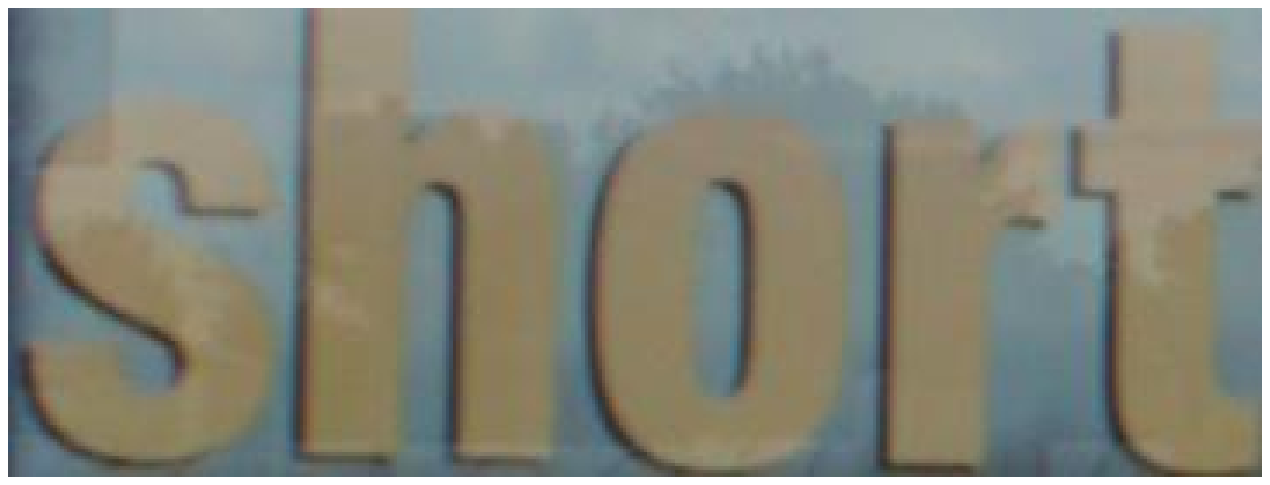}~\includegraphics[width=3.6cm]{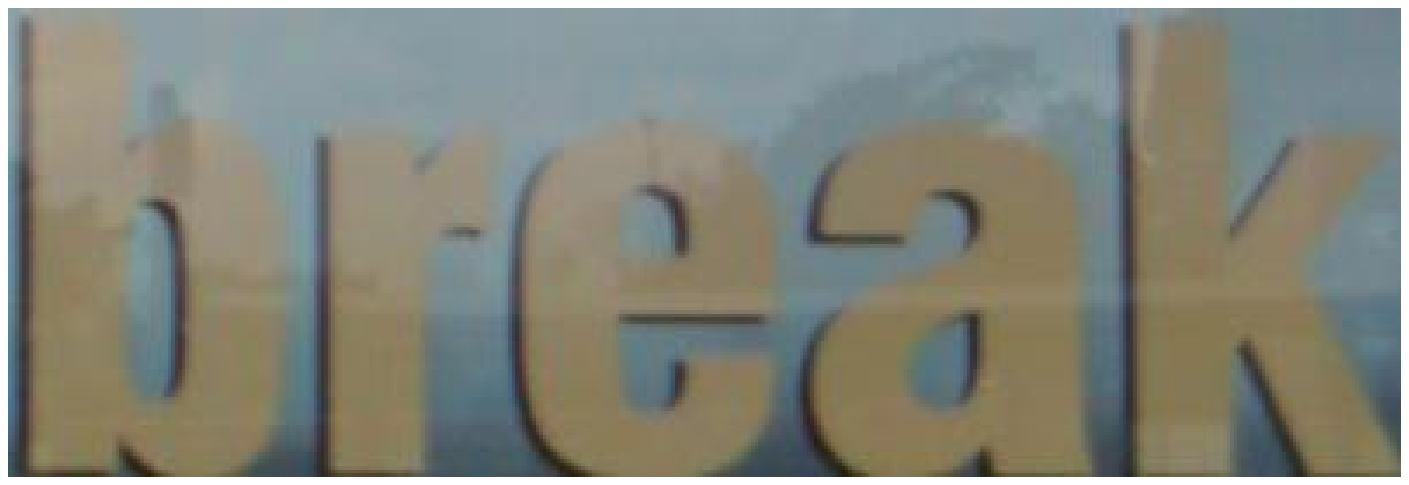}\\
\includegraphics[width=3.6cm]{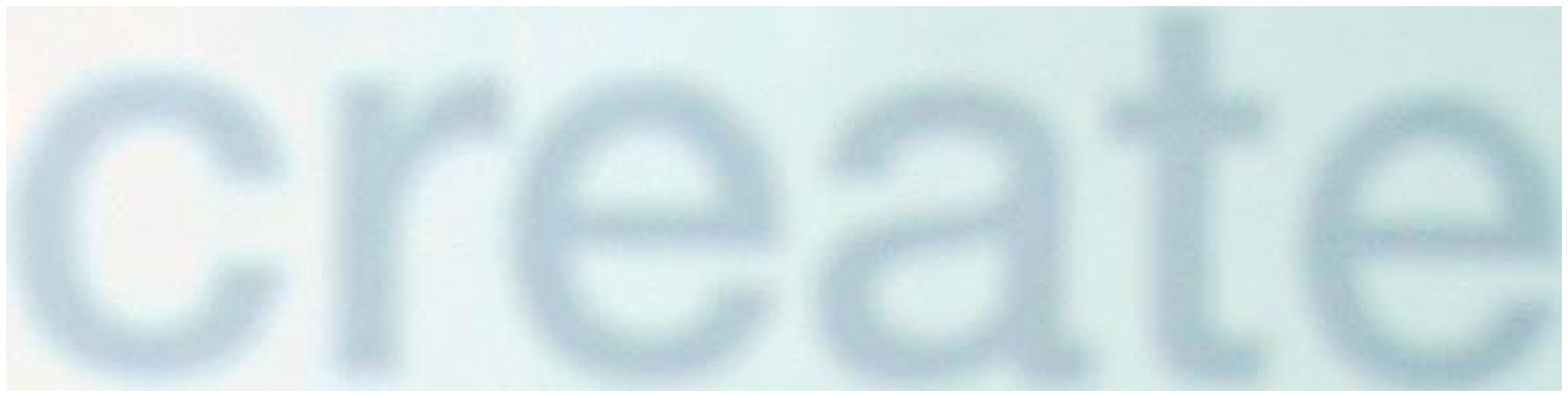}~\includegraphics[width=3.6cm]{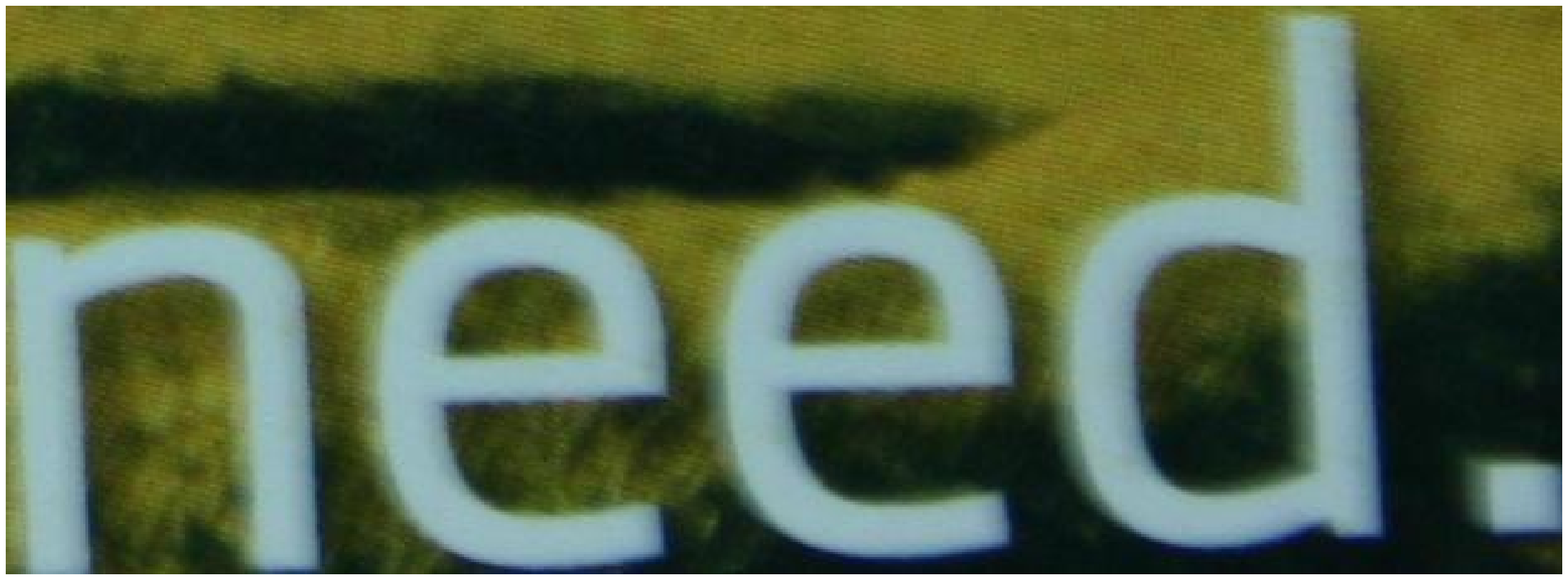}\\
\caption{Sample word images from ICDAR 2011 dataset \cite{asif}.}
\label{examples}
\end{figure}

There are several publicly available datasets for text localization \cite{tc11}. These datasets are known as IAPR TC11 Reading Systems-Datasets. One may assume that the bounding box information of a word is sufficient for any OCR to recognize. However we see that the best performing algorithm on ICDAR 2003 sample word image dataset (not the test set) has the word recognition rate of only around 52\%, without post processing using lexicon \cite{ananda}. Recently held ICDAR 2011, Robust Reading challenge 2 reports that the best word recognition rate is 41.2\% \cite{asif}. Figure \ref{examples} shows sample word images from this challenge.

Karatzas et. al initiated another robust reading challenge in ICDAR 2011 for born-digital images \cite{karatzas}. Born-digital images are formed by a software by overlaying text on an image. For the competition, these images were collected from web pages and email. Most words present in this dataset are oriented horizontally. The reason behind horizontal placement of text may be the simplicity involved in creating the born-digital image using standard softwares.

Low resolution of text and anti-aliasing are the main issues to be tackled in born-digital images, whereas  illumination changes and motion blur are difficult problems in the case of camera captured images. These issues indicate the complexity involved in processing born-digital and camera captured scenic images. 

An attempt for using top-down approach in word recognition can be observed in, sparse belief propagation with lexicon for word recongition by Weinmann et. al \cite{weinmann}. Similarly, Wang et. al use limited lexicon on Street view text (SVT) dataset \cite{wang10}. Both, Weinmann et. al and Wang et. al, use top-down approach for word recognition. They use an unsegmented character dataset to train a classifier. If the confidence of the character classifier is less, then top-down approach helps in classification using lexicon. Weinmann et. al use character level image annotation of the training data and textual features to classify the testing dataset. A limitation of this method is that it requires good quality character images with high resolution for training; else the classification will be erroneous. 

Wang et. al used Amazon's Mechanical Turk for annotation of SVT images. Bounding box was placed around the word spotted. The placement of these bounding boxes was not defined rigorously. The resulting irregularities in the word bounding boxes add additional complexity to the segmentation task and can be inferred from the low f-score reported. In the section on benchmark results of SVT dataset, we discuss as to how one can avoid this complexity.

\section{Annotation approach and tool}
Benchmarking is not a good idea, if annotation is not explicitly defined rigorously. We took five different datasets which have different definitions for bounding box and contain human errors while annotating bounding boxes for words. Our pixel level segmentation and annotation has been cross-checked thoroughly to reduce human errors to a minimum. 

A multi-script annotation toolkit for scenic text (MAST) was developed by MILE lab in 2011 \cite{mast}. It can be used to annotate scenic images. MAST has the facilities to annotate multiple scenic images or scenic word images. It has options for annotating multiple scripts. It has the additional capability for adding plug-ins with suitable layout for new scripts during annotation. It is publicly available for download\footnote{http://mile.ee.iisc.ernet.in/mile/download.html}.

MAST-CH, an enhanced version of character annotation tool kit has been recently developed by us. We discuss the differences between the two programs. MAST is designed to annotate scenic images with multiple word images with reasonably good resolution. Using seed points input by the user, the tool uses region growing and annotates at the pixel level, with a bounding box and text annotation for multiple scripts. On the other hand, MAST-CH handles a single word image at a time and annotates characters at the pixel level using multiple segmentation methods and user selection of output. It does not have provision to generate the text annotation for different scripts. Since some of the images in the datasets used contain low resolution images and truthed text is already available, we use MAST-CH toolkit to perform pixel level annotation . 
\begin{figure}[!t]
\centering
\includegraphics[width=9.0cm]{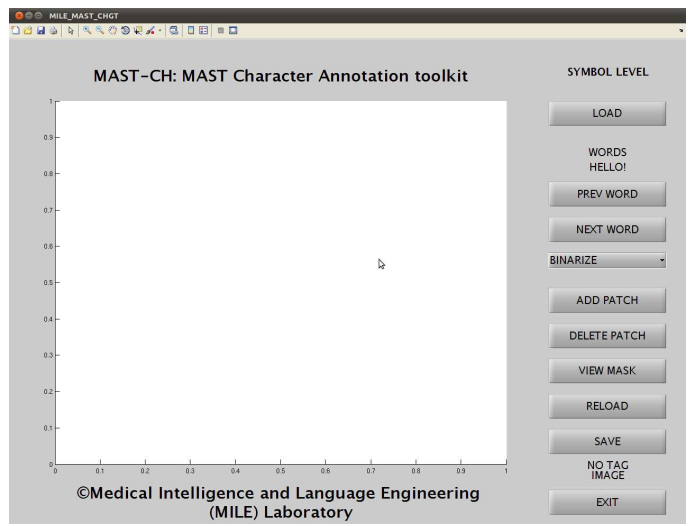}\\
\caption{Graphical user interface of the MAST-CH Annotation tool developed on MATLAB platform \cite{matlab}.}
\label{mgui}
\end{figure}

We have added new functionalities based on feedback from MILE lab project staff, who helped annotate the various datasets. A GUI of the tool kit with the buttons and a single window for image is shown in Figure \ref{mgui}. `LOAD' button enables us to load images from a particular directory. If a word image is highly degraded, and hence requires more time to annotate, it can be skipped. Those skipped images will not be tagged. `NEXT' and `PREV' buttons provide the user options for such skipping and going back during annotation, that help in rapid annotation of clean word images. `SAVE' button saves an annotated word image in .bmp format, also containing component ordering information and in .tiff format, containing colour map for individual components segmented). GUI also displays whether the currently loaded word image has already been tagged or not. `VIEW MASK' button overlays the obtained segmentation mask on the original word image.
 
\subsection{Enhancing speed of manual annotation}
In MAST, we segment words by region growing on the seeds placed by the user and then annotate the segmented words \cite{gonsalves}. Difficulty crops up when low resolution characters are to be annotated. To reduce the manual task and also to improve segmentation, we have removed the seed growing option. In place of it, we now use known segmentation algorithms.
 
For segmentation, we have provided a drop down button giving `BINARIZE' and `INVERT' options. The user can invoke the suitable option based on the relative colors of the foreground and background. Using multiple approaches, we create 16 different segmentation outputs. First, we split a colour image into the R, G, B planes and apply Otsu's threshold on each plane \cite{otsu}. We also convert the RGB image to HSV and CIE Lab space formats. Then, we split each of them into three planes and apply Otsu's threshold. In addition, we form three clusters using the RGB information directly and obtain the permutations for the clusters formed (each of the 3 clusters and union of any two clusters at a time). Finally, we apply robust automatic threshold selection algorithm on intensity of word image \cite{rats}. 

We display all of these segmentation results in another window and provide a manual keyboard input for the user to select one of the results. Once a user input is fed, a mask is generated and overlaid on the original image. By this way, we have removed manual seeding technique, which has improved the speed of segmentation task and reduced the fatigue of the annotators. If the mask generated has distinct or well separated characters, then the user can save the annotated result by clicking the  `SAVE' button. If none of the segmentation results are satisfactory, the user can choose `0' and thus no mask will be generated.
 
`RELOAD' button is used to load a saved mask and the corresponding original image. This is useful to examine annotated images. To minimize human errors, we cross-checked the annotated datasets three times.

\subsection{Use of polygons to refine segmentation}
A degraded image may not get segmented properly. This may be due to illumination changes, occlusion or low resolution of characters. To overcome these degradations, we provide a polygonal mask. These masks can be used to add parts of characters which are merged to the background or delete parts of the background that get added to a character. `ADD PATCH' button provides the option for adding pixels to the annotated mask in the polygonal format. `DELETE PATCH' button facilitates deletion of the background segmented as characters or splitting merged characters. When add or delete option is selected, we can place a single polygon at a given time. Mask will be modified based on the operation performed and the annotation tool asks whether to continue the same operation. If user chooses `yes', then the user can place another polygon to modify the annotated characters. If the choice is `no', then the tool exits this edit loop.

\begin{figure}[!b]
\centering
\includegraphics[width=2.7cm]{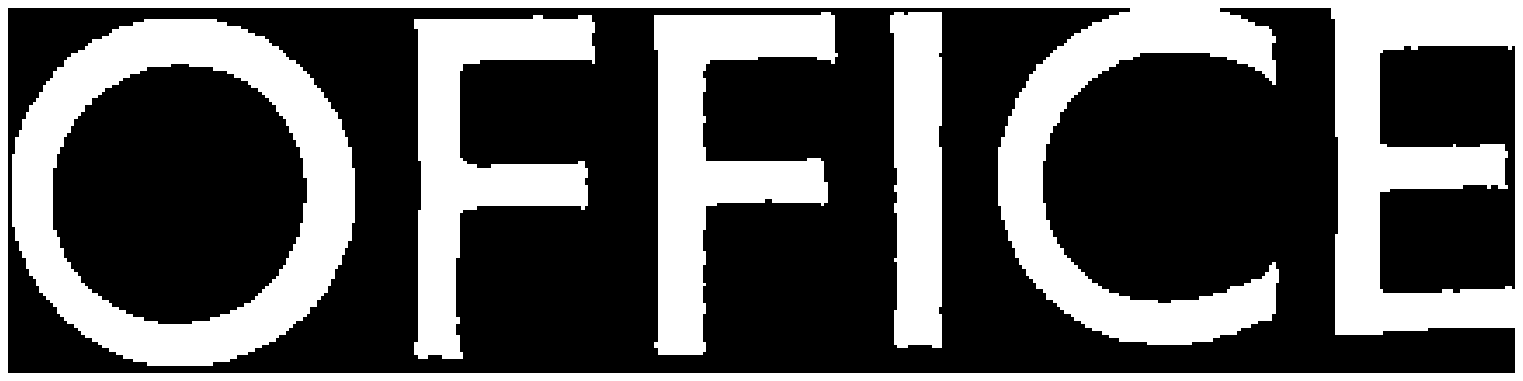}~\includegraphics[width=5.4cm]{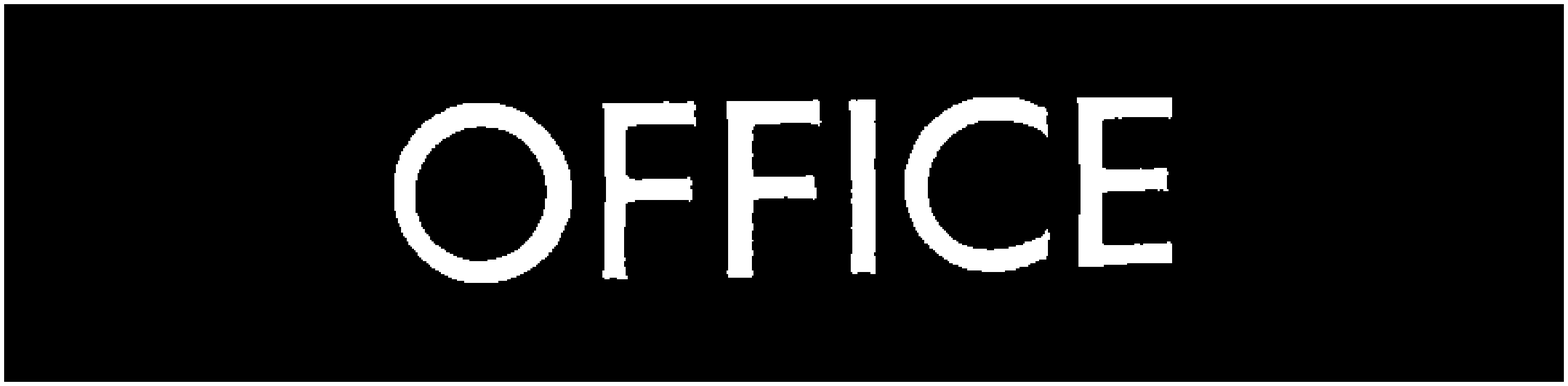}\\
\caption{Binarized image (a) without and (b) with preprecossing.}
\label{recpreprocess}
\end{figure}

\section{Recognition for benchmarking}
The pixel-level segmented images are fed to the recognition engine. Tesseract \cite{tesseract}, OmniPage \cite{omnipage}, Adobe Reader \cite{adobe} and Abbyy Fine Reader \cite{abbyy} are examples of readily available OCR engines. Any of these OCR softwares can be used to recognize the binarized word image. In our experiment, we use the trial version of Omnipage Professional 16 OCR for recognizing the word images. The recognition rate of the OCR on the above segmented word images is compared with the recognition results of the methods reported in the literature. In all these datasets, we can observe that the recognition rate on human segmented images is better than the rest. 

\subsection{Preprocessing to improve recognition}
Normally, any scanned document image contains top, left, right and bottom margins. However, as shown in Figure \ref{recpreprocess}(a), when we binarize a scene or a born-digital word image, margins do not exist since we have segmented at the word boundary.

In such cases, where characters touch the boundary, we observe difficulty in recognition with the OCR engine. To avoid this difficulty and also to provide margin in all directions, we add zero rows at the top and the bottom of the image, equal to half the original number of rows in the word image. Similarly, we pad zero columns on both the left and right sides of the word image. We refer these images as \emph {preprocessed} binarized images [see Figure \ref{recpreprocess}(b)]. Preprocessed binarized images are sent to the OCR for recognition. Recognition rates on binarized images are reported in the experimental section.

\section{Benchmark results on different databases}
We consider five word image datasets for experimentation. All these datasets are tagged using the annotation tool explained in Section 3. ICDAR, SVT and Born-digital word images have been annotated. Images with visually distinguishable boundaries between characters and background are tagged. Others have been ignored, since if a human cannot tag the text, we cannot expect an algorithm to either segment or recognize it. The annotated dataset is available for download from our MILE laboratory website\footnote{http://mile.ee.iisc.ernet.in/mile/download.html}. If any errors are observed, please report to the authors. 

These datasets cover different types of degradations except for motion blurs. All words in the dataset are tagged appropriately such that the visual distortion with respect to original image is minimum. In all the datasets, we have considered the testing set. We can improve the character segmentation using word images from the training set. We give below the recognition results for the five different datasets experimented upon.

\subsection{ICDAR 2003 dataset}
Robust Reading Competition was first conducted in ICDAR 2003 \cite{simon}. There were five entries for text localization and none for word recognition. Mishra et. al express the importance of binarization for word images and show 52\% as word recognition on sample dataset \cite{ananda}. This result explains that an equal importance should be given to word recognition. If we compare recognition rates of existing methods, this becomes more obvious. ICDAR 2003 Test dataset consists of 1110 word images, all of which are segmented by the authors. The word recognition rates are tabulated in Table \ref{icdar03table}.

Table \ref{icdar03table} shows a large gap in recognition rate between the preprocessed and non-processed images. This is because the low resolution text images are not recognized properly without proper margins formed by the background. Wang and Mishra et. al have used 829 images, a subset of ICDAR 2003 image dataset \cite{mishra12}. Hence, the reported result is averaged to the total number of images in the dataset. 

\begin{table}[!t]
\centering
\caption{Recognition rates on word images binarized by methods reported in the literature for ICDAR 2003 dataset.}
\label{icdar03table}
\begin{tabular}{|c|c|}
\hline
Algorithm & Word recognition rate\\
\hline
Human segmentation (Preprocessed) & {\bf 83.9}\\
\hline
Human segmentation & {\bf 68.0}\\
\hline
Mishra et. al \cite{mishra12} & 61.1\\
\hline
Wang et. al \cite{wang10} & 53.8\\
\hline
\end{tabular}
\end{table} 

\subsection{PAMI 2009 dataset}
The next two datasets we benchmarked were originally used to demonstrate the ability of top-down approach in word recognition. A lexicon provides information from the top layer to the middle layer during the classification stage. Using N-gram statistics derived from the limited lexicon formed by the datasets themselves, the authors show improvement in word recognition rate. 

Sign evaluation dataset was prepared by Weinmann et. al \cite{weinmann}. This dataset consists of 215 word images for analysis and was used to show top-down approach for character and word recognition. It consists of horizontally aligned characters only, except for one or two. The degradation in the images is also minimal. Hence the recognition rates reported by different methods are all close. The bench mark recognition rate on this dataset is the highest among all the datasets. The recognition rates for this dataset are tabulated in Table \ref{pami09table}.

\begin{table}[!t]
\centering
\caption{Comparison of recognition rates of existing methods on binarized word images from PAMI 2009 dataset.}
\label{pami09table}
\begin{tabular}{|c|c|}
\hline
Algorithm & Word recognition rate\\
\hline
Human segmentation (Preprocessed) & {\bf 89.3}\\
\hline
Human segmentation & {\bf 69.8}\\
\hline
Weinmann et. al (with lexicon) \cite{weinmann} & 86.1\\
\hline
Weinmann et. al (without lexicon) \cite{weinmann} & 79.1\\
\hline
\end{tabular}
\end{table}

\subsection{SVT 2010 dataset}
Wang et. al introduced street view text (SVT) dataset obtained as part of Google Street View project \cite{wang10}. Apart from the other degradations, these word images undergo motion blur. Severe motion blur and low resolution created difficulty for the authors in annotating a few word images (around 2\%).

This dataset consists only of name and location information of businesses. The bounding box tagged by Amazon's Mechanical Turk is not perfect. A rough bounding box is placed around the spotted word, which was listed by the Google search engine. This imprecise bounding box itself provides another layer of difficulty for locating the presence of text within the annotated bounding box. The erroneous tagging of bounding boxes has led to lower recognition rate using both open source and proprietary OCR engines. 

Mishra et. al show the improvement in recognition rate by top-down approach \cite{mishra12}. This result is biased, since it uses limited lexicon, built from the test word ground truth. With limited lexicon, you can hit upon the proper word more easily than with full lexicon, as discussed by Weinmann et. al\cite{weinmann}. Test word images are used to train the classifier for character segmentation, which boosts the recognition rate. If bounding boxes had been properly annotated, the advantage of lexicon would have resulted in a still higher recognition rate.  The recognition rates are tabulated in Table \ref{svt10table}. Table \ref{svt10table} shows that the top-down approach with limited lexicon has a higher recognition rate than other algorithms, but if proper segmentation is carried out, then the result would have been different.

\begin{table}[!b]
\centering
\caption{Recognition rate of binarized word image with existing methods for SVT 2010 dataset.}
\label{svt10table}
\begin{tabular}{|c|c|}
\hline
Algorithm & Word recognition rate\\
\hline
Human segmentation (Preprocessed) & {\bf 79.6}\\
\hline
Human segmentation & {\bf 74.8}\\
\hline
Mishra et. al & 73.3\\
\hline
Wang et. al & 56.0\\
\hline
Abbyy reader & 47.7\\
\hline
\end{tabular}
\end{table}

\begin{table}[!b]
\centering
\caption{Recognition rates of existing methods on binarized word images of BDI 2011 dataset.}
\label{bdi11table}
\begin{tabular}{|c|c|c|}
\hline
Algorithm & Edit distance measure & Word recognition rate\\
\hline
Human segmentation & 51.3 & {\bf 88.5}\\
(Preprocessed) & & \\
\hline
Human segmentation & 99.2 & {\bf 83.1}\\
\hline
Kumar et. al \cite{deepak12} & 108.7 & 82.9\\
\hline
Baseline \cite{karatzas} & 232.8 & 63.4\\
\hline
TH-OCR System \cite{thocr} & 189.9 & 61.5\\
\hline
\end{tabular}
\end{table} 

\subsection{Born-digital 2011 dataset}
Karatzas et. al initiated a new robust reading competition in 2011 \cite{karatzas}. This competition was based on email attached images and web images. These images are known as born-digital images, since they are created using software such as Adobe Photoshop \cite{adobe}, Gimp \cite{gimp} and Microsoft Paint \cite{paint}.

In these images, the text is placed by a user in an interactive fashion through a software. Hence, the fonts available in the system are only used to create the text pixels. When the recognizer uses a similar font, the recognition rate will be higher with born-digital images.

As mentioned earlier, the word images are obtained from the test images used in text localization and segmentation tasks of competition. A gap of four pixels exists from the boundary of text bounding box to provide the context of the image. This dataset had better word boundary definition than others.

This dataset consists of 918 word images. Only one participant competed in the competition. Abbyy fine reader was the baseline method used by the competition organizer to test the ability of algorithms. Since only TH-OCR algorithm competed in the competition and could not beat the baseline method, it was mentioned as honorary.  

The resolution of these word images are low compared to scenic word images. These images are affected by anti-aliasing. For their creation, born-digital word images use fonts that exist in the operating system. Due to low resolution and anti-aliasing, characters in the word images merge when a global threshold is applied. Kumar et. al applied power-law transform to remove the affect of anti-aliasing. By varying the $\gamma$ value in the power-law transform, they showed that the merged characters can be split \cite{deepak12}. Table \ref{bdi11table} compares the total edit distance and word recognition rate for born-digital image dataset. 

A new metric, edit distance measure, was introduced in this competition. This metric gives equal weights for addition and deletion of characters from the word. The calculated distance is normalized to that word. The normalized weights for all the words are added to form the edit distance measure for the full dataset. Total edit distance is tabulated in Tables \ref{bdi11table} and \ref{icdar11table}.

\begin{table}[!t]
\centering
\caption{Comparison of recognition rates on binarized word images using different methods for ICDAR 2011 dataset.}
\label{icdar11table}
\begin{tabular}{|c|c|c|}
\hline
Algorithm & Edit distance measure & Word recognition rate\\
\hline
Human segmentation & 60.1 & {\bf 86.7}\\
(Preprocessed) & & \\
\hline
TH-OCR System \cite{thocr} & 176.2 & 41.2\\
\hline
KAIST AIPR System \cite{kaist} & 318.5 & 35.6\\
\hline
Neumann's method \cite{neumann} & 429.8 & 33.1\\
\hline
\end{tabular}
\end{table} 

\subsection{ICDAR 2011 dataset}
This dataset, a subset of ICDAR 2003 dataset, was used in ICDAR 2011 Robust reading challenge Task 2. It consists of 716 word images. In this dataset, a few additions have been made and repeated words from the ICDAR 2003 dataset have been removed. Those removed images are from the scene images and were not considered either in the testing or training of ICDAR 2011 competition.

The recognition rates of existing methods are shown in Table \ref{icdar11table}. We can observe that the recognition rate has improved with respect to ICDAR 2003 dataset. There were three entries for word recognition competition in ICDAR 2011: Robust Reading Competition Challenge 2. So, we have included this dataset for discussion.

Here, we could not access the recognition rate of non-preprocessed binarized image. Due to polarity reversal of some images by OCR itself, this resulted in erroneous text. The reason is that the bounding boxes specified are tight. Word images in the test dataset do not have any additional pixels around the word boundary, as discussed in Born-digital 2011 dataset.

\section{Discussion}
We have prepared pixel level annotation for five word image datasets. We took this huge task of annotation, in order to show that segmentation of word images is important to recognize characters/words. Even though top-down analysis is useful in improving the recognition rate on specific datasets using a limited lexicon, it is not practical in real world situation. Weinmann et. al showed that the recognized word rate reduces with full lexicon.

\begin{figure}[!t]
\centering
\includegraphics[width=4.5cm]{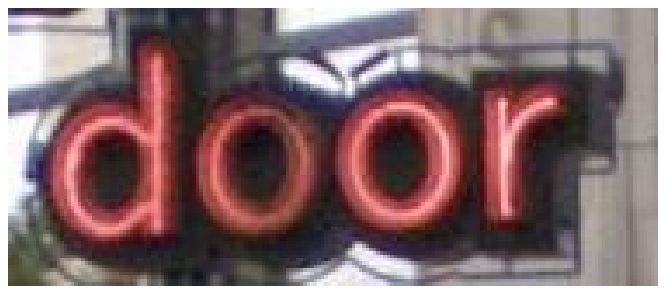}~\includegraphics[width=4.5cm]{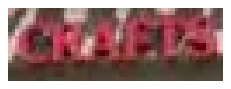}\\
\includegraphics[width=4.5cm]{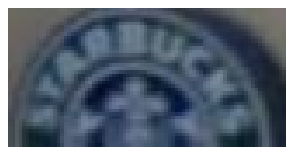}~\includegraphics[width=4.5cm]{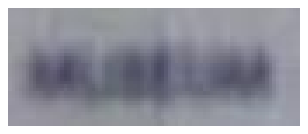}\\
\caption{Street view images with different degradations (a) Improper bounding box (b) Low resolution (c) Curved text and (d) Motion blur}
\label{svtimages}
\end{figure}

Around 85\% of word recognition is achieved with manual segmentation. Thus, if we provide more importance to proper segmentation of characters countering all degradations, we can improve the recognition. Here, all word images were segmented in such a way that individual components in the segmented image can be properly recognized or classified by a classifier or an OCR engine.

In this paper, we infer that if we train dataset specific classifier with annotated word images, then we can use the training dataset for word recognition. Skewed or curved words in the images can be classified better by a custom-built classifier than an OCR engine. We can observe that in street view dataset, the recognition rate of words is often poor due to skew or curvy nature of words. Figure \ref{svtimages} shows sample images from street view dataset with different degradations. Hence, the trained classifier will help in improving the recognition rate. In the case of skewed or curved words, a trained classifier is less affected and with minimal processing, we can improve the recognition rate. Standard OCR engines do not provide this functionality. Also we use individual test characters segmented to measure stroke width of the characters, which helps in improving the segmentation as a top-down approach.

\section{Conclusion}
We have completed the annotation of five standard databases. We have made the annotated datasets publicly available for download from our MILE website\footnote{http://mile.ee.iisc.ernet.in/mile/download.html}. Any one can download and test them using any OCR engine. The recognition rate differs across OCR engines and also with the versions. From all the tabulated results, it is evident that we need to improve the segmentation algorithms to get better word recognition. Our approach indicates the requirement for good segmentation, since it is the major part of the bottom-up approach. We can use lexicon information to improve the recognition rate reported. The validity of our good segmentation can be indirectly seen from the achieved word recognition rates.

\section*{Acknowledgment}
We express our heartfelt thanks to Shanti Devaraj, Shanti S and Saraswathi S, who were involved in the segmentation of word images. Improvement in UI for word image annotation was possible only from their feedback. The annotation work has minimal errors. The credit goes to them for careful annotation and committed interaction with the authors. Finally, without them, our dream to annotate all the images at the pixel level would not have been accomplished.   

\end{document}